# ROMAN URDU OPINION MINING SYSTEM (RUOMiS)

Misbah Daud[1], Rafiullah Khan[2], Mohibullah[3] and Aitazaz Daud[4]

[1, 2, 3] CS/IT Department, IBMS, University of Agriculture, Peshawar, Pakistan
misbahdaud@gmail.com  rafiyz@gmail.com  mrmohibkhan@gmail.com
[4] Department of Computer Science, CUSIT, Peshawar, Pakistan
zazikhan1@gmail.com

## ABSTRACT

*Convincing a customer is always considered as a challenging task in every business. But when it comes to online business, this task becomes even more difficult. Online retailers try everything possible to gain the trust of the customer. One of the solutions is to provide an area for existing users to leave their comments. This service can effectively develop the trust of the customer however normally the customer comments about the product in their native language using Roman script. If there are hundreds of comments this makes difficulty even for the native customers to make a buying decision. This research proposes a system which extracts the comments posted in Roman Urdu, translate them, find their polarity and then gives us the rating of the product. This rating will help the native and non-native customers to make buying decision efficiently from the comments posted in Roman Urdu.*

## 1. INTRODUCTION

In every business, customer's feedback and opinions caries an important value for number of reasons. These feedbacks or customer opinions are gift to the business because these can help them to improve products, offerings and services and most of the times companies get new ideas. However there is another angle to see the opinions, "customer convincing the *customer*". Yes it's true whilst of detailed description and reviews of an experts, there will always be those who want more [1].

Trust is a huge issue in online business. A big number of customers still hesitate to buy products from e-stores. Their fear from shopping at online store may be driven by security issues, credit card information theft, goods quality, shipment procedure or may be concerns over the reliability of the seller. To develop the trust of the customer, one such solution is to provide the facility of to the customer to leave their comments. New customer normally either follow the current trend or looking for an independent review. Thanks to the technological advancement in the web, a commenting system provide a facility to the visitors to share their experience with other visitors and make them customer from visitor. Infect nowadays social platforms have become more actives in these sorts of activities. An integration of e-store with social platforms attract more visitors to the store by just posting the reviews over the customer's wall.

Opinion mining is an evolving area of Information retrieval. This area is a combination of Data mining and Natural language processing techniques. Customer opinion mining is a process to find user's insight about different products [2]. These customers' observations are found in the form of comments posted by the customers. Opinion or comment is user generated content and normally posted in unstructured free-text form.

Users normally comment in their native language which is understandable by the very few people who speaks that language. The significance of those comments is very little. What if we go one step further and make these comments useful for those users who don't understand any other languages but English? The crux of this research is to help the non-native customer to get rating of the products from the comments posted in Roman Urdu. Urdu is the national language of Pakistan and spoken in other five Indian countries. D+ue to complex morphology and some technological limitation, Urdu script is not so common. Roman Urdu is a term used for the Urdu language written in Roman script [3]. Similarly Romanagari is portmanteau of two words Roman and Devanagari [4]. Romanagari is a term used for Hindi language written in Roman script. In Pakistan and India people normally use Roman Urdu and/or Romanagari for posting the comments. Urdu and Hindi are same languages however the difference between them is writing script and influence of other language [5]. Urdu is influenced by Persian, Arabic and Turkish. Its writing style is in Arabic while Hindi is written in Devanagari script [4]. However both languages are much more similar and can be understandable by the people living in subcontinent (if it is written in Roman script). Let for example take a phrase posted in Roman Urdu "Iss mobile ka camera acha ha". Its English translation will be like "The camera of this mobile is good". Similarly "Fig 1" shows the same comment in Urdu script. The same phrase "Iss mobile ka camera acha ha" is fully understandable in Hindi. Our focus in this paper will be Roman Urdu.

اس موبائل کا کیمرہ اچھا ہے۔

Fig 1: Comment in Urdu script.

This paper proposes an application RUOMiS (Roman Urdu Opinion Miner System), an automatic opinion mining system that mine and translate the Roman-Urdu and/or Romanagari reviews and provide the rating of the products based on users comments. To help the non-Urdu speaking users in selection of product.

The paper is divided into five sections. First and second section contains the introduction and related work respectively. In third section the architecture of the RUOMiS and methodology is briefly described. In forth section results of the initial experiments are discussed, the effectiveness of the system using information retrieval matrices and last section states the conclusion and future directions.

## 2. RELATED WORK

In 2004 Liu and Hu made one of the early attempts on mining and summarizing customer reviews. They proposed a system an opinion summarization system which uses NLP (Natural Language Processing) linguistics parser [2]. That parser tags the parts of speech for each word in the sentence. They used association miner algorithm for mining frequent features on noun/noun phrases. For the classification of positive and negative opinions, adjectives words were analysed and WordNet [6] dictionary was used for the orientation of words. They also used the same system and attempt for the mining the opinions which carries some specific features of the interested product [7]. Another system with the name OPINE was proposed by Popescu et al [8]. OPINE is unsupervised information extraction systems which is capable of identifying the features of the products and the opinions related to that, determine the polarity and then rank the opinions according to strengths. Wang et al proposed a web based product review and customer opinion summarization system with the name of SumView [9]. SumView is using the Feature-based weighted Non-Negative Matrix Factorization (FNMF) algorithm for classifying sentence into relevant features cluster. It provides summary of information contained review documents by selecting the most representative reviews sentences for each extracted product feature.

All the system discussed above perform opinion mining and a sort of sentiment analysis. However their work revolves around English language. De-cheng and Tian-fang proposed a framework for topic identification and features extraction from the opinions and reviewed posted in Chinese language [10]. El-Halees proposed a systems which extract the user opinion from the Arabic text [11]. His proposed system uses three methods i.e. lexicon based method, machine learning method and k-nearest model for better performance. Abbasi et al worked on the sentiment analysis of the opinions posted in English or Arabic. They used Entropy Weighted Genetic Algorithm ( EWGA) for the assessment of key features [12]. Almas and Ahmad used computational linguistics for English, Arabic and Urdu. However their work is purely based on the financial news data and describe a method of mining some special terms from the data which they called it local grammar [13]. Syed et al proposed an approach of sentiment analysis using adjective phrase [14]. Javed and Afzal proposed lexicon based bi-lingual sentiment analyses system that analyze the tweet of users written in two language i.e. English and Roman-Urdu. They used two type of lexicon for sentiment analysis. For English language they used SentiStrength lexicon [15] and for Roman-Urdu, they manually developed a lexicon, as there is no other such lexicon is available for Roman-Urdu [16].

## 3. ARCHITECTURE OF RUOMiS

The purpose of this research is to provide a facility to the non-Urdu speaking customers to get benefit from the comment posted in Roman Urdu. For experiment purpose a website whatmobile[1] is selected. This web site detailed information of cell phones of well-known cell phone companies in Pakistan. This site also provide facility of comments over each cell phone page. Most of the comments are written in Roman Urdu however some users also comments in English too. The architecture of the RUOMiS is given the Figure 2.

The system consists of four main steps. Crawling of site, Translation of Roman-Urdu reviews into English language, Identification of the opinion polarity and giving rating in graphical form. However each step have some sub steps. The detail of each step and sub step is given below.

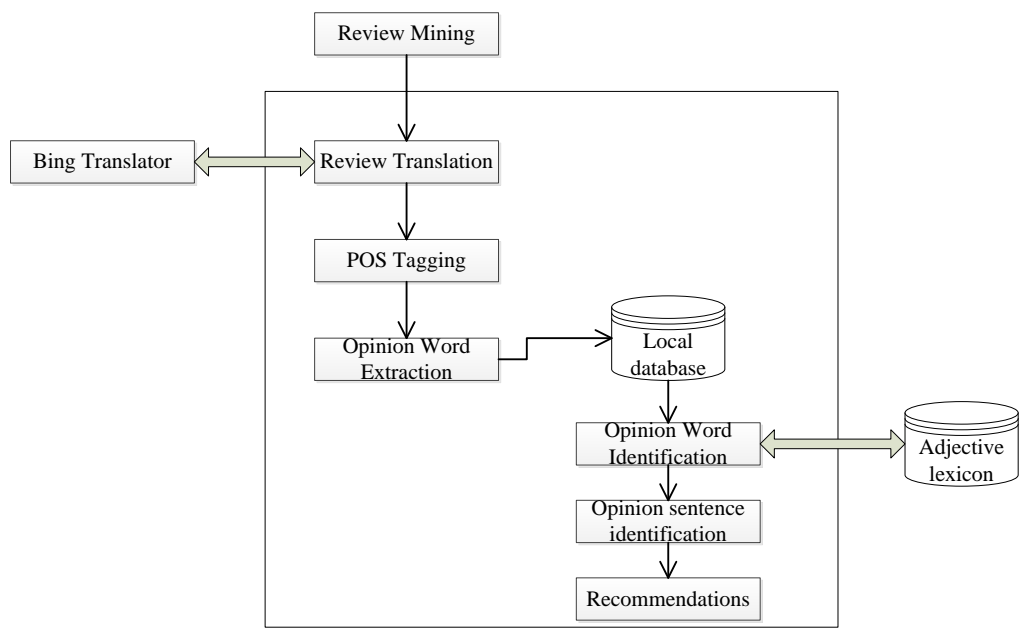

Figure2: RUOMiS Architecture.

### 3.1 Review Mining and Review Translation

The review miner will first crawls all the reviews from the input URL and will store them into local temporary storage. After fetching the data will be handed to the Review Translation module. The Review

---

1 http://www.whatmobile.com.pk/

Translator module will query the comments over Microsoft Bing Translation Service using API. After getting the translated text back, next step is to analyse and tag it.

### 3.2 Parts of Speech Tagging

As the identification of the opinion polarity is based on the adjective used in the sentence, it is important here to identify the adjectives in the sentences. Part of Speech Tagging is used for that identification purpose. SharpNLP [17] is used for POS tagging. It's a C sharp port of Java OpenNLP tools. OpenNLP is a library of machine learning based toolkit [18]. It have the built-in facilities of text processing like tokenization, segmentation of sentences, PoS tagging, parsing chunking and many others. It's a very powerful tool even it can be used to extract triplets from the text [19]. In this case the SharpNLP perform all the pre-processing of data like tokenization, sentence segmentation, and part-of-speech tagging. The following sentence shows a sentence with POS tags. "The pictures are very clear." Will become "The/DT pictures/NNS are/VBP very/RB clear/JJ." Where DT stands for Determiner, NNS for Plural Noun, VBP for Verb and third person and present and JJ stands for Adjective.

### 3.3 Opinion Words Extraction

Opinion word expresses the subjectivity of sentence that whether it is positive, negative or neutral. These words help in finding the orientation of sentence. Mostly people show their opinion about product using opinion words. Adjectives are describing words that classify its object or noun in sentence. Adjective words are extracted from review database using their adjective tag applied by the POS tagger. For example: "The pictures are very clear." In this sentence, word clear is adjective describing the picture quality.

### 3.4 Opinion Sentence Identification

After identification of opinion word, the semantic orientation (Positive or negative) of each sentence are determined. Any sentence, which contains opinion word, considered as opinion sentence. To investigate the optimism and pessimism of the opinion sentence, an opinion word dictionary is designed manually. Initially it consists of 200 positive and negative adjectives. The semantics of opinion word are identified by comparing each opinion word with local opinion dictionary. If the opinion word is positive, it is assigned to the positive category to that sentence. Opinion sentence mainly depends on the semantic orientation of the opinion words. If the sentence doesn't belong to either of the category, it will be considered as neutral.

### 3.5 Rating / Recommendation

After the identification of the opinion polarity, the system make a summary of all the comments and give the output rating of the product. The rating of the product then can be integrated in the webpage specific product. The rating of the product could be represented in figures, pie chart or bar chart.

## 4. RESULTS AND DISCUSSION

Three cell phones are selected for experiment purpose, the proposed system have analysed 1620 comments. The comments are classified in three categories Positive comments, Negative comments and Neutral comments. Positive comments are those comments in which user admire some feature of the product. Negative comments are those comments in which user report some complaint against the product. While Neutral comments are those in which user just replicate some of the feature of the product or a user just reply to other user comments which carries no polarity like comments shown in Table 1.

However this should also be noted that there were some comments which was not relevant to the product at all. Those comments were posted by either some machine or by some user those comments contain some sort of advertisements. Those comments are considered as neutral comments. Table 1 contains different types of neutral comments observed during the crawling. Table 1 contains some of the neutral comments that were found while crawling. Table 1 also shows that weather those comments are relevant to the product and also weather those comments are posted by some machine or human being.

Table 1: Neutral Comments

| Comments | Relevancy to Product | Type |
|---|---|---|
| Lol thanx… | Relevant | User Comment |
| Do it contain skype??? | Relevant | User Comment |
| I need a backup tool | Noise | User Comment |
| www.xyz.com is the website you need for change | Noise | Machine Comment and Advertisement |
| I have nokia xyz in good condition. Only serious buyers can call. | Noise | User Comment and Advertisement |
| I wanna buy this cell...plzz somebody tell me whats exactly release date is...? | Relevant | User Comment |
| I m selling BB curve 8900. Good condition. Demand is 4500/- Not Negotiable. rwp isb contact 92 334 xxx 0000. | Noise | User Comment and Advertisement |
| agr kise k pass ya mobile hai good condition mai tou mujy contact kare 0333x0xx1xx argent | Relevant | User Comment and Advertisement |
| hy mara pas nokia 108 ha 10 month warinti new condichn final demand 2700 full box lahore 03xx 8xx5xx9 | Noise | User Comment |
| Aslam o alaikum.frndz mjhe xperia x8 ki orignal battery chahye kisi dost ne sale krni ho to plz cntct me 0xx-2xx9xx8x frm lahore. | Noise | User Comment |

Table 2: Results of products comments polarity

|  | Product # 1 | | | Product # 2 | | | Product # 3 | | |
|---|---|---|---|---|---|---|---|---|---|
|  | Positive | Negative | Neutral | Positive | Negative | Neutral | Positive | Negative | Neutral |
| RUOMiS | 164 | 52 | 324 | 210 | 60 | 270 | 153 | 65 | 322 |
| Manually | 51 | 24 | 465 | 38 | 27 | 475 | 31 | 20 | 489 |

Table 3: Total result of all products.

|  | Positive | | Negative | | Neutral | |
|---|---|---|---|---|---|---|
|  | Comments | Percentage | Comments | Percentage | Comments | Percentage |
| RUOMiS | 527 | 32.5 % | 177 | 10.9 % | 916 | 56.5 % |
| Manually | 120 | 7.4 % | 71 | 4.4 % | 1429 | 88.2 % |
| Deviation | 0.251 | | 0.065 | | 0.317 | |
| **Total Average Deviation from Original** | = | 0.211 | | | | |

The comments posted on three different products are consider and analysed. 540 comments for each products were taken which make the total of 1620. The comments are also analysed manually for comparison purpose. The results of all three products are provided in the in Table II. Similarly Table III contain total or all three products.

Out of all three selected products, RUOMiS reported that 527 comments were positive comments while the actually there were 120 positive comments. Similarly RUOMiS reported 177 negative comments while the actual figure was 77 and RUOMiS reported 916 neutral comments while the actual neutral comments were 1429. On the average results of RUOMiS are 21.1 % deviated from the original results.

Table 4: contingency table

| Orientation | Number of Comments |
|---|---|
| True Positive (TP) | 191 |
| False Positive (FP) | 513 |
| False Negative (FN) | 0 |
| True Negative (TN) | 916 |

The Precision, Recall and the F-measure of RUOMiS are calculated. Table IV classify the comments in four classes: true positive, true negative, false positive and false negative. True positive are the comments that were correctly identified by the RUOMiS as either positive or negative. False positive are comments which are actually not neutral but RUOMiS categorize them as neutral. False negative are the comments which RUOMiS falsely reported as neutral but that was not actually neutral. And true negative are comments which are correctly identified as neutral comments. The following equations were used to calculate the precision, recall and F-measure of the system.

$$\text{Precision} = TP / TP + FP \quad \text{...........................................} (1)$$

$$\text{Recall} = TP / TP + FN \quad \text{...........................................} (2)$$

$$F = 2* \text{Precision}*\text{Recall} / \text{Precision} + \text{Recall} \quad \text{...........................} (3)$$

The precision of the RUOMiS was 0.271, recall was 1.0 and F-measure was 0.427. Where precision is number of all actual positive and negative comments retrieved by the RUOMiS divided by the number all positive and negative retrieved comments. While recall is the number of all actual positive and negative comments retrieved by the RUOMiS divided by the number all positive and negative comments in the whole batch. So according to the result the recall of the relevant comments is 1.0 which means all positive and negative comments are selected however RUOMiS also selected about 21.1% of neutral comments as positive of negative comment falsely.

## 5. CONCLUSIONS AND FUTURE DIRECTIONS

The objective of this research was to help non-Urdu speaking customers to get benefit from the comments posted about some product in Roman Urdu. As the Romanagari is also supported here this makes the tool more versatile. This research proposed an architecture that uses Natural Language Processing technique to find the polarity of the opinion. In addition to that an own lexicon are developed which contain positive and negative adjectives. The plan is to make that lexicon available for public and give a facility to the public to add new words in the database, however the lexicon is maintain in the SQL database in which data cannot be stored semantically. Thanks to WordNet an online semantically maintained. WordNet is extensible, semantically well maintained and easily accessible lexicon.

The results of experiment are satisfactory as the recall of relevant results is 100% however RUOMiS categorized about 21.1% falsely. We have identified some of the reasons the first and biggest reason is noise in data. Almost 80% of the neutral comments were about advertisement of their used goods or request for the used goods. Every comment of this nature carries the condition of the goods like "Excellent

Condition" or "Good Condition" and other similar adjectives. These sort of comments are falsely classified in positive category as shown in the table 3, where total number of positive comments are 120 but RUOMiS reported 527 and similarly total number of neutral comments 1429 while RUOMiS reported 916. And negative comments difference little i-e 6.5% as compere positive and neutral comments difference which are 25.1% and 31.7% respectively. Second reason is the algorithms of the "Opinion Word Identification" and "Opinion Sentence Identification" modules. As the opinion polarity is decided based on the Adjectives, if comment is composed on two or more sentences the algorithm consider it two different comments. There is another reason related to the translation service. A Microsoft's Translator is used which is considered as one of the powerful translation solution however it cannot be accurate 100% due to spelling and grammatical mistakes of users.

Overall the results are satisfactory. A notable results with precision of 27.1% is achieved and could be enhanced. The further enhancement is always possible by introducing some method of noise detection. For this purpose semantic solutions of will be better option. A Semantic dictionary of noisy data could play vital role in detecting noise. WordNet could also be used for this purpose. Tuning of the algorithms could be beneficial.

**Authors**

**Misbah Daud** received BS Degree from the Institute of Business and Management Sciences (IBMS), The University of Agriculture Peshawar Pakistan Currently perusing MS degree in IT from same University. Area of Interest is Data Mining and Information Retrieval.

**Rafiullah Khan**, Lecturer Institute of Business and Management Sciences (IBMS), The University of Agriculture Peshawar Pakistan received his BS degree in Computer Science from University of Peshawar in 2007 and MS degree in Information Technology from the Institute of Management Science (IM|Sciences) Peshawar, Pakistan in 2010. Currently, he is perusing Ph.D. from the Department of Computer Science of the Muhammad Ali Jinnah University, Islamabad. His fields of specialization are Semantic Computing, Information Retrieval and Computer Based Communication Networks.

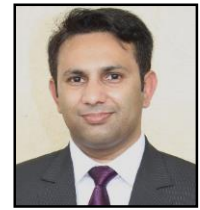

**Mohibullah**, Lecturer Institute of Business and Management Sciences (IBMS), The University of Agriculture Peshawar Pakistan received M.Sc. degree in Data Networks and Security from the Birmingham City University, United Kingdom. Currently, he is perusing Ph.D. from the Department of Computer Science of the Muhammad Ali Jinnah University, Islamabad. His fields of specialization are Semantic Computing, Information Retrieval and Data Mining.

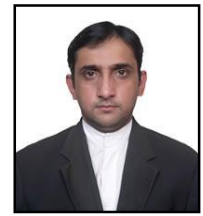

**Aitazaz Daud**, received BS (Software Engineering) Degree from City University of sciences and Information Technology, Peshawar Pakistan. He is working as programmer at Broadway Solutions.